\documentclass{article}
\usepackage[T1]{fontenc}
\usepackage[utf8]{inputenc}
\usepackage{amsfonts}
\usepackage{float}
\usepackage{hyperref}
\usepackage{mathtools}
\usepackage{spconf,amsmath,graphicx}
\usepackage[backend=biber,style=ieee,natbib=false,mincrossrefs=1000]{biblatex}
\usepackage{fancyhdr}
\AtBeginDocument{\toggletrue{blx@useprefix}}
\AtBeginBibliography{\togglefalse{blx@useprefix}}

\newcommand{\R}{\mathbb{R}}
\DeclarePairedDelimiter{\norm}{\lVert}{\rVert}

\allowdisplaybreaks[2]
\title{Towards a Semantic Perceptual Image Metric}

\name{Troy Chinen, Johannes Ballé, Chunhui Gu, Sung Jin Hwang, Sergey Ioffe, Nick Johnston,}
\secondlinename{Thomas Leung, David Minnen, Sean O'Malley, Charles Rosenberg\sthanks{Work done while at Google; now at Pinterest, Inc., San Francisco, CA 94107, USA}, George Toderici}
\address{Google AI Perception\\Mountain View, CA 94043, USA}

\begin{document}

\maketitle

\begin{abstract}
We present a full reference, perceptual image metric based on VGG-16, an artificial neural network trained on object classification.  We fit the metric to a new database based on 140k unique images annotated with ground truth by human raters who received minimal instruction.  The resulting metric shows competitive performance on TID 2013, a database widely used to assess image quality assessments methods.  More interestingly, it shows strong responses to objects potentially carrying semantic relevance such as faces and text, which we demonstrate using a visualization technique and ablation experiments. In effect, the metric appears to model a higher influence of semantic context on judgments, which we observe particularly in untrained raters. As the vast majority of users of image processing systems are unfamiliar with Image Quality Assessment (IQA) tasks, these findings may have significant impact on real-world applications of perceptual metrics.
\end{abstract}


\fancypagestyle{firststyle}
{
   \fancyhf{}
   \lfoot{\textcopyright 2018 IEEE}
   \setcounter{page}{1}
   \cfoot{\thepage}
   \rfoot{ICIP 2018}
}

\thispagestyle{firststyle}

\fancyhf{}
\renewcommand{\headrulewidth}{0pt}
\setcounter{page}{1}
\cfoot{\thepage}

\section{Introduction}
\label{sec:introduction}
IQA is a difficult task even for human raters, as it requires mapping an extremely large space of possible images and distortions onto a single number. Raters tend to show significant variability in their responses, in particular to IQA tasks where the distortions are well above the human discrimination threshold. For instance, raters may differ in how they weight the severity of different types of distortions (such as distortions in the luminance vs. chrominance channels), or in how they spatially integrate the presence of several distorted regions in an image. Instruction and training is commonly used to control variability. For instance, raters may be discouraged from letting image semantics influence their decisions, or the first $N$ ratings may be discarded to allow them to adapt to the task. Trained raters typically develop a stable integration scheme for the types of distortions they are presented with, and may even adapt to individual distortions or test images -- they “know where to look”. Although this may help to reduce variability, it may not help to accurately represent perceptual judgments made by users of image processing systems “in the wild”. We have observed that minimally instructed or trained raters tend to focus on semantically relevant objects in the scene.

\begin{figure}
\includegraphics[width=\columnwidth]{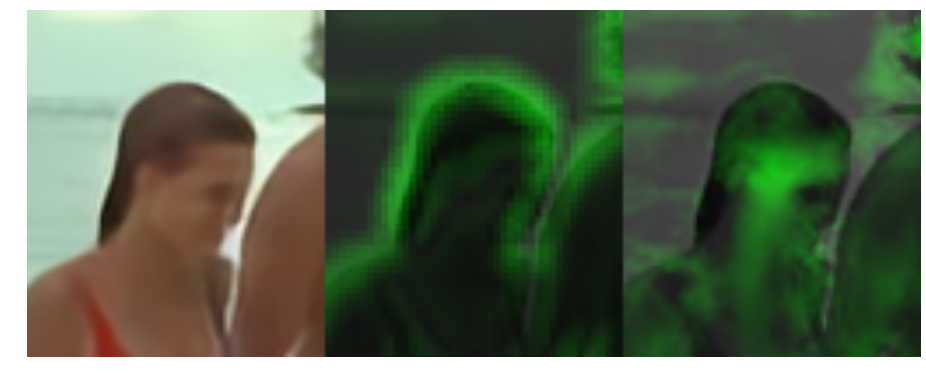}
\caption{Perceptual metrics can be trained to respect high-level semantic information such as faces.  (a) the distorted image, (b) its perceived distortion under a contrast-based metric and, (c) the same for the proposed perceptual metric, indicating that distortions to a face are more objectionable than distortions to edges.  Higher green intensity means higher predicted distortion.}
\label{fig:teaser}
\end{figure}

This implies that the performance of IQA models in real-world scenarios could be enhanced by giving them access to higher-order image features. Many existing image metrics such as SSIM~\cite{WaBoShSi2004}, MS-SSIM~\cite{WaSiBo2003}, PSNR-HVS~\cite{EgAsPoLuBa2006}, PSNR-HVS-M~\cite{PoSiEgCaAs2007}, FSIM~\cite{ZaZaMoZh2011}, Butteraugli~\cite{Al2016}, HaarPSI~\cite{ReBoKuWi2018} are designed to model specific documented aspects of the human visual system (HVS), such as contrast or color sensitivity, but are unable to access features that would be able to discern object classes, for instance. One avenue for exploring this idea is to use artificial neural networks (ANNs) that have been trained for object classification tasks. It has been shown that their feature spaces can rival that of the inferior temporal cortex in primates~\cite{CaHoYaPiAr14}, and researchers have used them successfully for image processing applications like texture synthesis~\cite{JoAlFe2016} and super resolution~\cite{LeThHuCaAi2016}.

In this paper, we present preliminary results exploring the use of pre-trained ANNs for modeling quality judgments. We collected a dataset of ground truth data for an IQA task from untrained raters. We then designed an image quality index with 10 parameters based on VGG-16, a pre-trained ANN, and fit it to the data. While our metric performs competitively on an existing IQA database, it outperforms all other metrics on another distinct dataset we collected from untrained raters. It appears to do so by utilizing higher-order image features inherent to VGG-16.

\section{Collection of ground truth}
\label{sec:training-data}
Reference images were generated from 10MP JPEG images which were downsampled by a factor of 4 to reduce compression artifacts. For each of these 140k reference images, two distorted images were created independently by applying a random sequence of distortions (of random length), sampled with replacement from 8 distortion types: compression artifacts from JPEG and a variant of an ANN-based compression method~\cite{ToViJoHwMi17}, two types of Gaussian noise, blur, posterization, gamma correction, and contrast rescaling. For example, an image could be distorted by blur($\sigma=4.5$), then JPEG(Q=60), and finally gamma($\gamma=1.9$). Subsequently, each image was cropped to a single random patch of $224\times 224$ pixels to reduce the variability of rater responses by limiting the influence of different spatial integration strategies.

Inspired by~\cite{PoJiIeLuEg2015,MaToMa2012,ShKaPh2015}, a protocol similar to two-alternative forced choice (2AFC) was employed, except that raters had the additional option to declare UNSURE.  2AFC is better suited for uninformed raters than mean opinion score (MOS), because it avoids the need for calibration.  Raters were presented image \emph{triplets}: a reference image O, and 2 distorted images A and B. The task was to decide which of A or B is most similar to the reference. The full-size reference image was presented along with the cropped versions for context.

We took care to minimize raters' familiarity with image processing, distortion types, etc. We avoided any training sessions or materials which might bias raters in terms of adapting to or recognizing particular distortions, or which might discourage them from applying their own rationale for discrimination.  Raters were instructed to maintain a distance of 0.5m to the 92dpi 24'' monitor in the remote facility.  Aspects such as lighting and session duration were uncontrolled for, as in real world conditions.  We employed 200 raters in an attempt to sample the space of preferences; each triplet received 5 ratings from this pool. The final collection of patches contained 700k ratings with an average rating time of 3.49s per triplet.

\section{Proposed metric and model fitting}
\label{sec:perceptual-metric-model}
\label{sec:training}
We define our full reference perceptual image metric as
\begin{equation}
\label{eq:metric}
f(x, y) = \sum_i w_i \norm{\phi_i(x) - \phi_i(y)}_1 = W \Phi(x, y),
\end{equation}
where $x$ and $y$ are images, $w_i \in \R$ are model parameters, and $\phi_i$ is a vector containing the responses of the $i$th layer of VGG-16  (directly after the rectified linear units).  The layers are the 5 conv and 5 pool layers of VGG-16. Other choices, such as using only lower or only higher layers, did not perform as well as the full set of convolutional layers.  The parameters of the VGG-16 model were used as pre-trained on ImageNet~\cite{RuDeSuKrSa2015} using $224\times 224$ images over 1000 classes.  During training, the $W$ are optimized while the $\Phi$, which are determined by the pre-trained VGG-16 weights, remain fixed.

The goal is for $f$ to respect a distance-like property, i.e., larger values correspond to larger images differences:
\begin{equation} 
\label{eq:ranking}
f(o,a) < f(o,b) \iff a \triangleright b,
\end{equation}
where $o$, $a$, $b$ are original image, and two distorted versions, respectively, and $a \triangleright b$ means ``$a$ is judged by humans to be closer to the original than $b$''.  Using \eqref{eq:metric}, the above condition above can be rewritten as $WX_{o,a,b} < 0$ where $X_{o,a,b}$ is defined as $\Phi(o,a) - \Phi(o,b)$.  We omit the dependence of $X_{o,a,b}$ on the images $o, a, b$ going forward.

Now consider a binary classification problem where a feature vector $X$ has target 1 if $b \triangleright a$ and 0 otherwise.  We use logistic regression to train such a classifier. The output of the logistic regression is the decision function $F(X) = g(WX)$ where $g(x) = (1+e^{-x})^{-1}$, which we train to represent the probability that $b \triangleright a$. This is accomplished by maximizing $E_{a,b: b \triangleright a} [\log g(WX)]$ with $L_2$ regularization on $W$. We do not include a bias term inside the decision function since we wish to have $F(X) = 1-F(-X)$.  The desired weights are $W$.

\section{Experimental Results}
\begin{figure}
 \centering
 \includegraphics[width=\columnwidth]{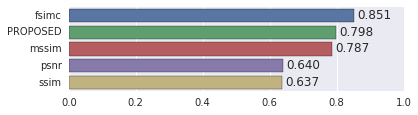}
 \caption{The proposed metric's performance on TID2013 vs. a sample of other methods.  Other data reproduced from \cite{PoJiIeLuEg2015}.}
 \label{fig:tid-spearman}
\end{figure}

\begin{figure}
 \centering
 \includegraphics[width=\columnwidth]{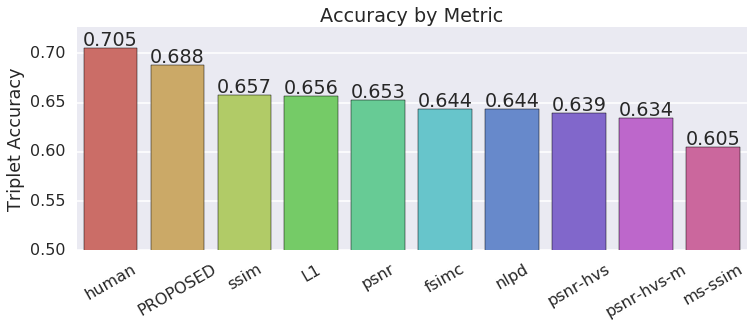}
 \caption{Accuracy in predicting human pairwise rankings in a compression dataset.  Proposed metric approaches human accuracy.}
 \label{fig:compression-accuracy}
\end{figure}

\begin{figure}
 \centering
 \includegraphics[width=\columnwidth]{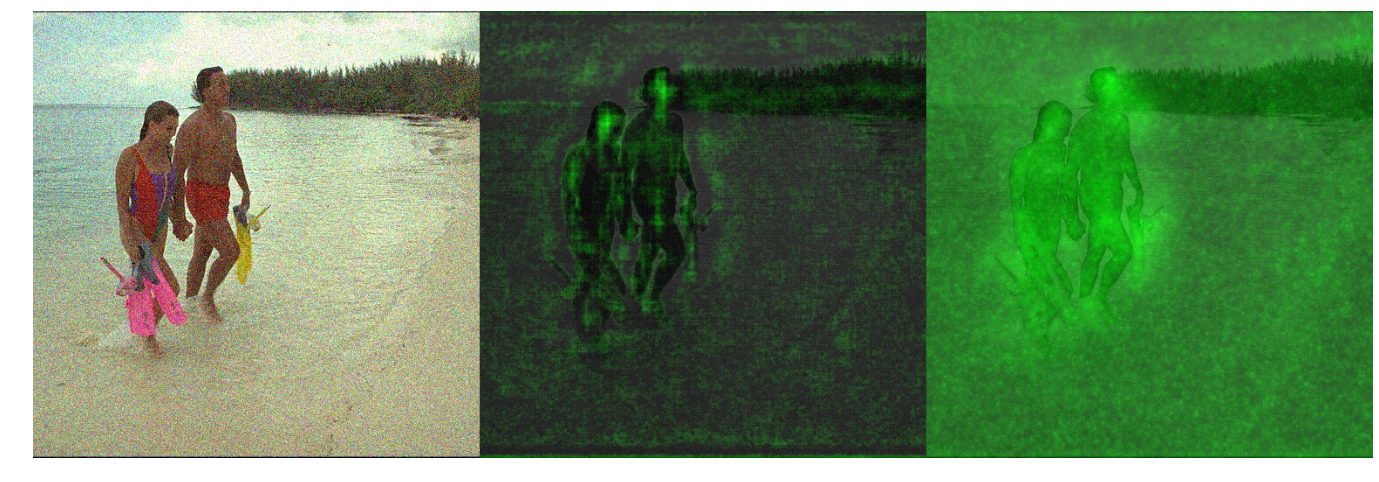}
% \hfill
 \caption{Trained vs not-trained proposed metric.  (Left) the distorted image, (middle) the proposed metric, and (right) the proposed metric with training turned off (i.e. $w_i = 1$).}
 \label{fig:notrain}
\end{figure}

\begin{figure}
\includegraphics[width=\columnwidth]{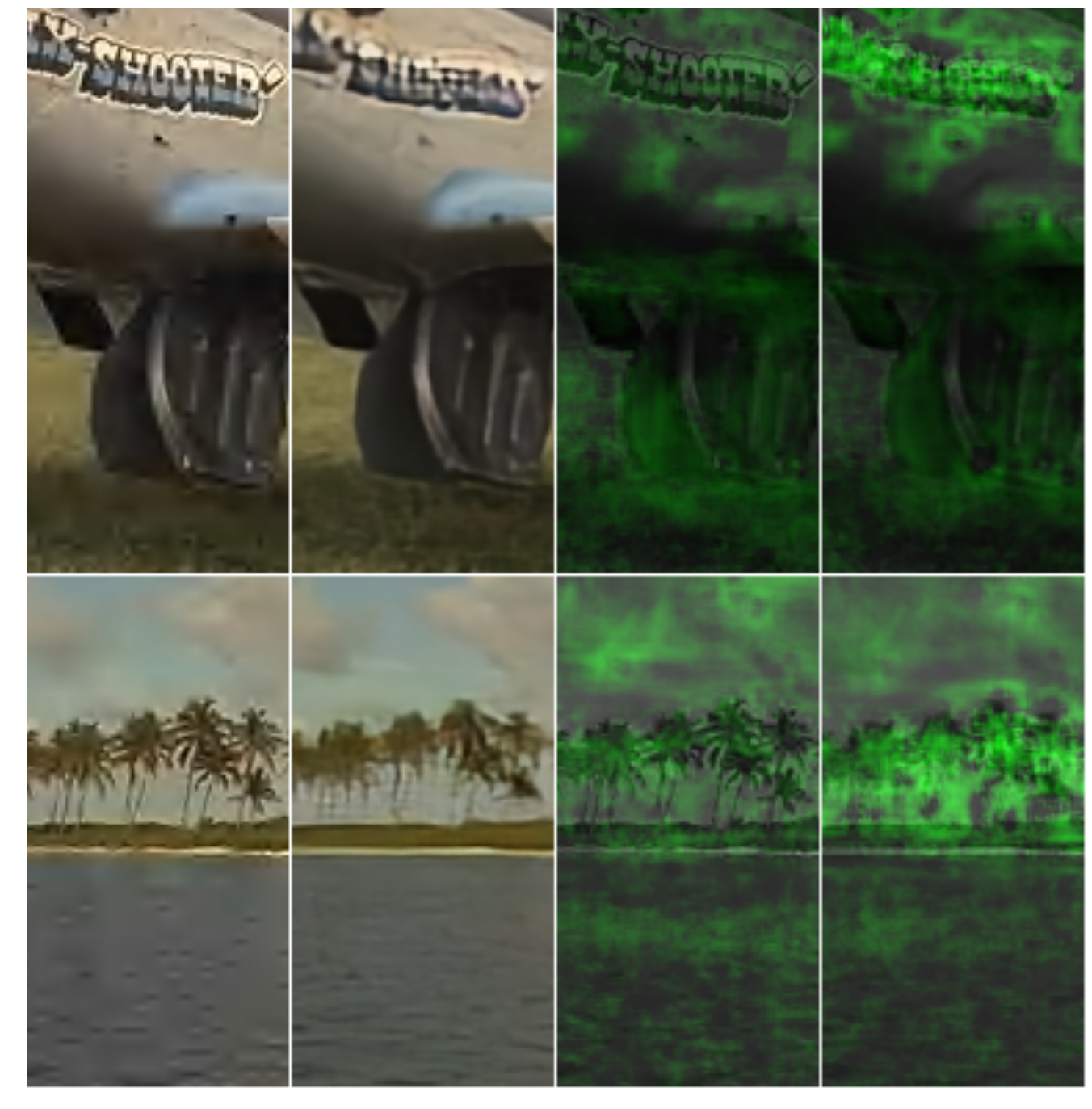}
\caption{Predicting human ratings.  Columns are (a) - (d) from left to right; (a) \& (b) are distorted images and (c) \& (d) are the corresponding proposed heatmaps.  In row 1, despite artifacts in large areas like grass, 10/10 raters chose (a) and the heatmap predicts that the text distortion is most objectionable.  In row 2, 9/10 raters chose (a) despite wavelet noise in the sky and water; the heatmap predicts that the palm tree distortion is most visible.  In both rows the overall proposed metric scalar value predicts (a) to be chosen while FSIM predicts (b).}
 \label{fig:raters-good}
\end{figure}

\begin{figure}
 \centering
 \includegraphics[width=\columnwidth]{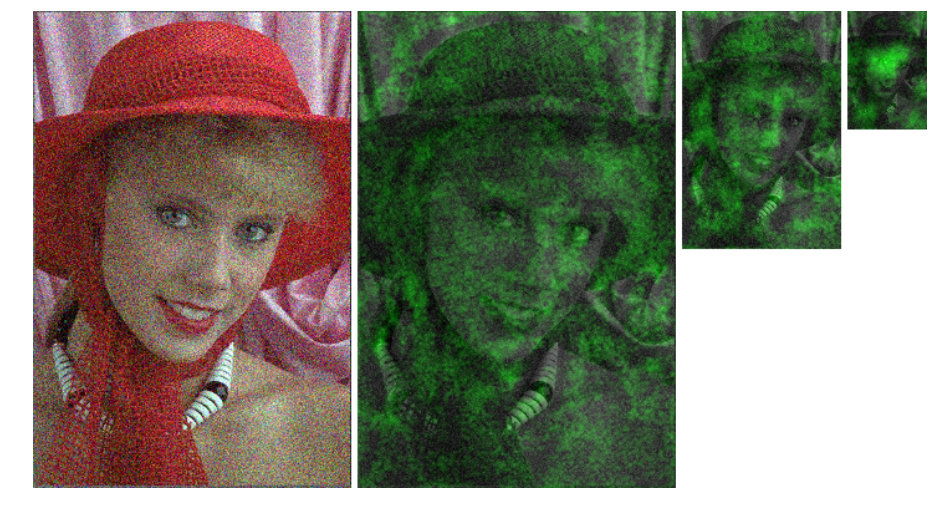}
 \caption{
 Failure case.  The proposed metric responds to objects at the scale presented to the network at training time.  (a) the distorted image, (b) - (d) the metric response on downsampled versions of the image (noise is added after downsampling).  The strong response to a face is seen only in the smallest image; there, the face is approximately 90 pixels high which is what would be expected in the VGG training data.}
 \label{fig:scale}
\end{figure}

\label{sec:experimental-results}
\label{sec:quantitative-performance}
The proposed metric was evaluated on TID2013~\cite{PoJiIeLuEg2015} where it achieves SROCC and KROCC of 0.798 and 0.615 respectively, placing it in the top 5 of 16 performers on both measures. \autoref{fig:tid-spearman} shows SROCC with a representative selection of metrics.  Note that the TID2013 images are disjoint from those in the training set.

We also constructed a compression dataset using the 24 Kodak~\cite{Kodak1991} images distorted by 8 lossy compression algorithms at compression rates from 0.125 to 1 bits per pixel.  Ground truth was established as with the training data, except with 10 ratings per triplet, and randomized crops to $512\times 512$ pixels. \autoref{fig:compression-accuracy} shows the accuracy (the ratio of correctly predicted triplet rankings to total triplets) of several full reference metrics on this compression dataset.  The proposed metric outperforms the other metrics and approaches human performance, which is the average inter-rater agreement.

\begin{figure}
 \centering
 \includegraphics[width=\columnwidth]{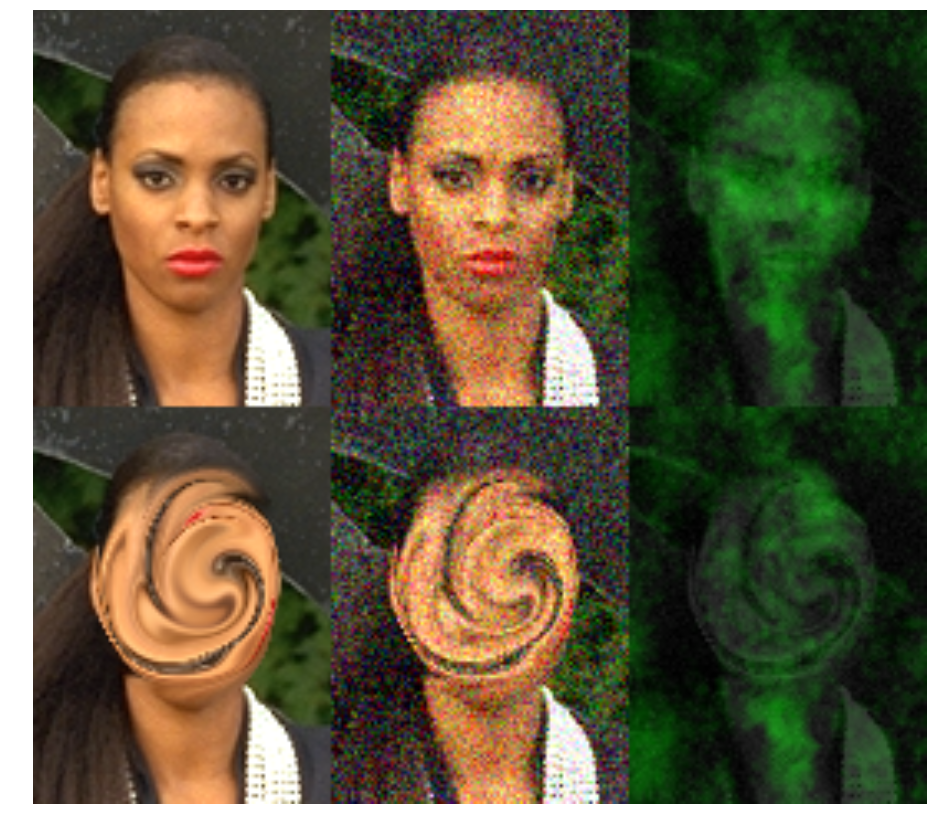}
 \caption{The proposed metric responds to object level features. Columns: reference, reference + Gaussian noise (fixed random seed), metric response.  In row 2 the face in the reference image is destroyed.  Despite the face pixel values being similar, the metric responds much weaker to the same noise distortion, indicating that some objects are treated preferentially.}
 \label{fig:noface}
\end{figure}

\label{sec:qualitative-performance}
The performance discrepancy between the two datasets may be attributed to the different sets of distortions in TID~2013 compared to our training data, or to the different evaluation methodologies -- we use logistic regression on raw triplets and rank order metrics, whereas TID~2013 applied some postprocessing before publishing the data, and correlation to the MOS. However, in light of the discrepancy we observe between experts and minimally instructed raters, the reason may also lie in the influence of image semantics. Seeking anecdotal explanation, we recruited and interviewed a local volunteer. We asked her to rate according to the same protocol, and provide explanations without any guidance or prompting. Her self-reported rationale clarified that she recognized distortions, but consciously discounted ones she considered less semantically important (e.g., ``If you took a picture of this [house on a cliff above the ocean], you're taking a picture of the house, not the sky, so I don't care if the sky looks bad.'').

\label{sec:gaussian-dataset}
To analyze the discrepancy, we developed a technique to visualize the image regions which contributed most to the predicted judgment. We defined a simple “heatmap” version of the proposed metric by omitting the spatial summation in \eqref{eq:metric}, producing a value at each pixel.  We constructed a similar heatmap visualization for HaarPSI as an example of an existing metric modeling contrast sensitivity. In the figures, higher intensity of green indicates more predicted distortion.  We observe that the heatmaps generated by our metric are qualitatively different from the ones generated by HaarPSI in that they tend to respond much less predictably to simple features such as edges (figure~\ref{fig:teaser}). In particular, we find that there are often strong responses to faces and other objects that could have semantic relevance, such as text (figures~\ref{fig:notrain}, \ref{fig:raters-good}, \ref{fig:scale}). This is a surprising result, given that there are only 10 model parameters and that the training data consists of random patches, thus probably containing relatively few faces.

The pronounced response to faces is remarkably robust: when we only apply pseudo-random Gaussian noise ($\sigma=30$, identical seeds) to rule out any interactions caused by complex signal-dependent distortions such as compression artifacts, and compare identical images with and without pixel scrambling (figure~\ref{fig:noface}), scrambled faces elicit a much weaker response. Put simply, a noisy face is predicted to be more objectionable than a noisy non-face. Similarly, if we don't fit the model parameters to ground truth data and simply set $w_i=1$, the responses are much less specific (figure~\ref{fig:notrain}).

To further assess differences between our metric and existing work, we selected image triplets from the compression database where predictions differed between our metric and FSIM. \autoref{fig:raters-good} shows examples where the proposed metric agreed with human raters. In instances where our metric failed to predict human responses correctly, we observed that it often appears to be particularly sensitive to smaller background objects, discounting large foreground objects. This may be caused by a scale dependence in VGG-16: it was trained on images of $224\times 224$ pixels, and hence the feature detectors it provides are small in scale compared to the Kodak image. \autoref{fig:scale} illustrates such a failure case.

\section{Conclusion}
\label{sec:conclusion}
We collected a ground truth dataset consisting of 700k human judgments and used it to fit a full reference image quality model based on VGG-16, an ANN pre-trained for object classification. It performs competitively on TID 2013, and outperforms existing metrics on a dataset with compression artifacts collected from minimally instructed raters, approaching human performance. Our analysis seems to indicate that it relies on higher-order image features generated by VGG-16, such as object detectors, to predict the human judgments.

Our work is related to existing work in using ANNs for full-reference IQA, such as \cite{BoMaMuWiSa16,KiLe17}. However, we are unaware of other publications that do this in the context of minimizing rater instruction and training. Our work is also conceptually related to saliency maps, which are designed to capture bottom-up processes of attention in the human visual system. In the context of region-of-interest video compression, for instance, researchers have used measures of saliency to modulate image quality metrics, or the compression algorithm directly, in order to introduce a semantic weighting of the content (e.g., \cite{It04}). A compelling aspect of our results is that the proposed metric seems to use an internal measure of saliency that was inferred from the ground truth data, implying that an explicit model of saliency may be unnecessary.

Many questions remain as to how perceptual metrics with a better model of image semantics can be designed, and what other factors contribute to the discrepancy we observed between the datasets. Collecting ground truth data with minimal instruction and training is more challenging than from raters who are familiar with IQA tasks, because it generally increases variability in their responses. However, our results suggest that a large part of this variability may in fact be systematic, and could be explained by models that have access to higher-order image features, such as ANNs. A future direction of research may be to develop improved protocols for collecting ground truth that maximize the raters' freedom to apply semantic judgments, yet minimize inter- and intra-subject variability. Depending on the application, it may also be desirable to design metrics with varying levels of semantic modeling, which should be reflected in the protocol.

\section{REFERENCES}
\printbibliography[heading=none]

\end{document}